\documentclass{article}
\usepackage{spconf,amsmath,graphicx,amssymb}
\usepackage{enumitem}
\usepackage{booktabs} 
\usepackage{float}
\usepackage{hyperref}
\usepackage{algorithm}
\usepackage{xcolor}
\usepackage{algpseudocode}
\setlist{nosep, leftmargin=14pt}

\title{Synergizing Deep Learning and Biological Heuristics \\ for Extreme Long-Tail White Blood Cell Classification}

\name{
Duc T. Nguyen$^{1,2,3}$ \quad
Hoang-Long Nguyen$^{1,2,3}$ \quad
Huy-Hieu Pham$^{1,2,3,*}$\sthanks{Corresponding author: Huy-Hieu Pham (hieu.ph@vinuni.edu.vn).}
}

\address{
$^{1}$ College of Engineering \& Computer Science, VinUniversity, Hanoi, Vietnam   \\
$^{2}$ VinUni-Illinois Smart Health Center, VinUniversity, Hanoi, Vietnam\\
$^{3}$ The Computer Vision and Medical AI Lab, VinUniversity, Hanoi, Vietnam}

\begin{document}
%\ninept
\maketitle

% --- ABSTRACT ---
\begin{abstract}

Automated white blood cell (WBC) classification is essential for leukemia screening but remains challenged by extreme class imbalance, long-tail distributions, and domain shift, leading deep models to overfit dominant classes and fail on rare subtypes. We propose a hybrid framework for rare-class generalization that integrates a generative Pix2Pix-based restoration module for artifact removal, a Swin Transformer ensemble with MedSigLIP contrastive embeddings for robust representation learning, and a biologically-inspired refinement step using geometric spikiness and Mahalanobis-based morphological constraints to recover out-of-distribution predictions. Evaluated on the WBCBench 2026 challenge, our method achieves a Macro-F1 of 0.77139 on the private leaderboard, demonstrating strong performance under severe imbalance and highlighting the value of incorporating biological priors into deep learning for hematological image analysis. The code is available at \url{https://github.com/trongduc-nguyen/WBCBench2026}

\end{abstract}


% --- SECTION 1: INTRODUCTION ---
\section{INTRODUCTION}
\label{sec:intro}
Automated white blood cell (WBC) classification is crucial for leukemia screening, yet deep learning (DL) models severely underperform on real-world, long-tailed clinical datasets \cite{asghar2024classification, yang2022survey}. In hematology, this extreme class imbalance is compounded by high intra-class variation, subtle inter-class morphological similarities, and synthetic domain shifts, leading to systematic misclassification of rare pathological subtypes \cite{wang2023intra}. 

To address these compounding challenges, we propose a robust three-stage hybrid framework integrating generative domain restoration, hierarchical representation learning, and expert-driven morphological refinement. Our method explicitly mitigates severe data scarcity and domain shifts while enhancing rare-class semantic separability, securing a top-tier ranking on the official WBCBench 2026 Challenge \cite{wbcbench2026}.
% --- SECTION 2: PROPOSED METHODOLOGY ---
\section{PROPOSED METHODOLOGY}
\label{sec:method}

\subsection{Dataset and Distribution Analysis}
\label{ssec:data}

\begin{figure}[h]
  \centering
  \includegraphics[width=1.0\linewidth]{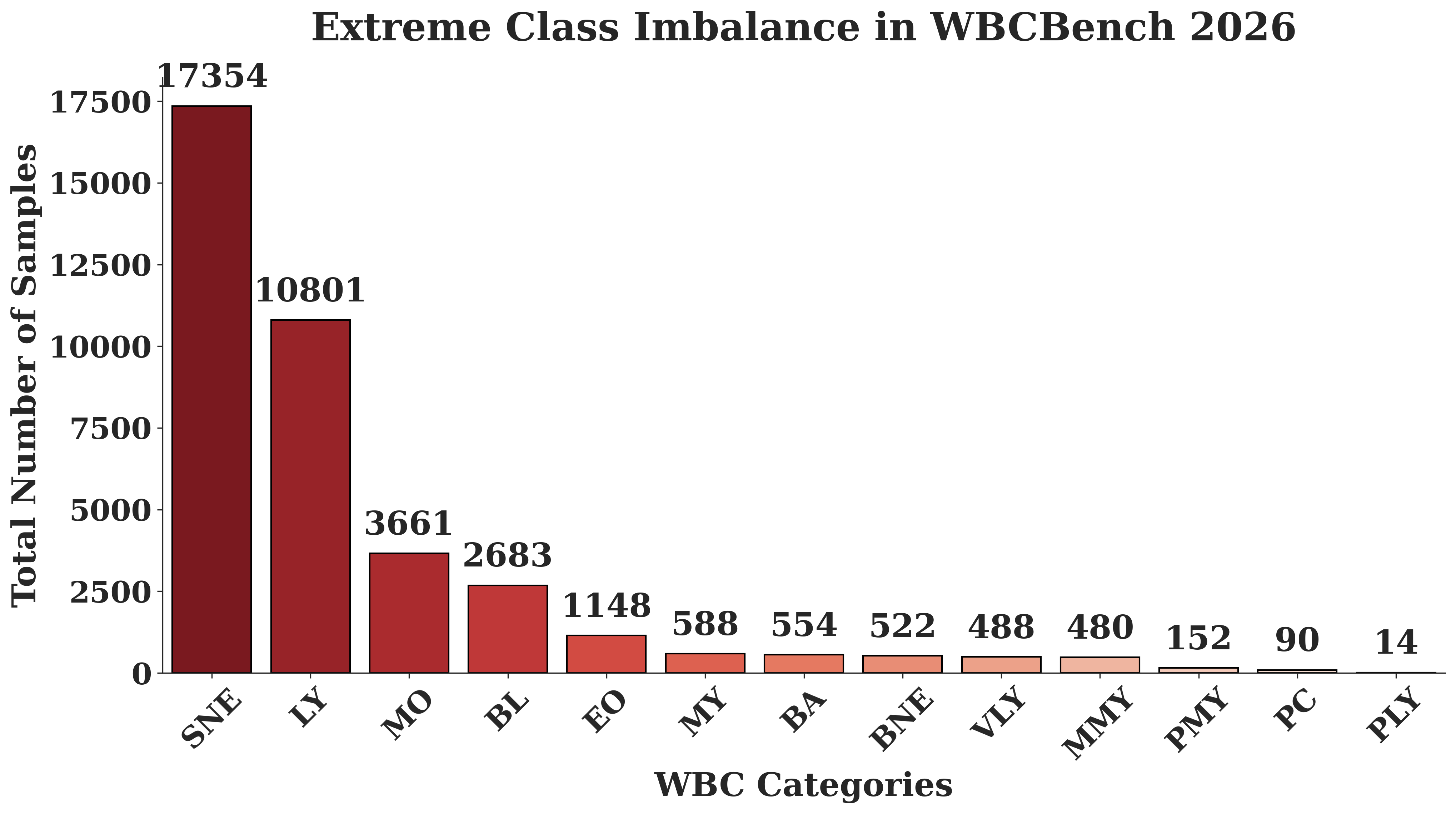}
  \caption{Class distribution of the WBCBench 2026 dataset.}
  \label{fig:distribution}
\end{figure}

The WBCBench 2026 dataset comprises 55,012 peripheral blood smear images, categorized into 13 fine-grained white blood cell (WBC) classes by expert hematopathologists. As visualized in Fig. \ref{fig:distribution}, the dataset exhibits an extreme long-tail distribution that faithfully replicates real-world clinical prevalence.
The majority class, Segmented Neutrophils (SNE), dominates the training set with 17,354 samples. In stark contrast, rare pathological classes such as Plasma Cells (PC) and Prolymphocytes (PLY) are represented by only 90 and 14 samples, respectively. The 1,200-fold class imbalance induces an extreme long-tailed recognition setting, where minority classes operate under severe sample scarcity.
\subsection{Preprocessing and GAN Restoration}
\label{ssec:preprocess}
\begin{figure*}[t]
  \centering
  \includegraphics[width=0.9\textwidth]{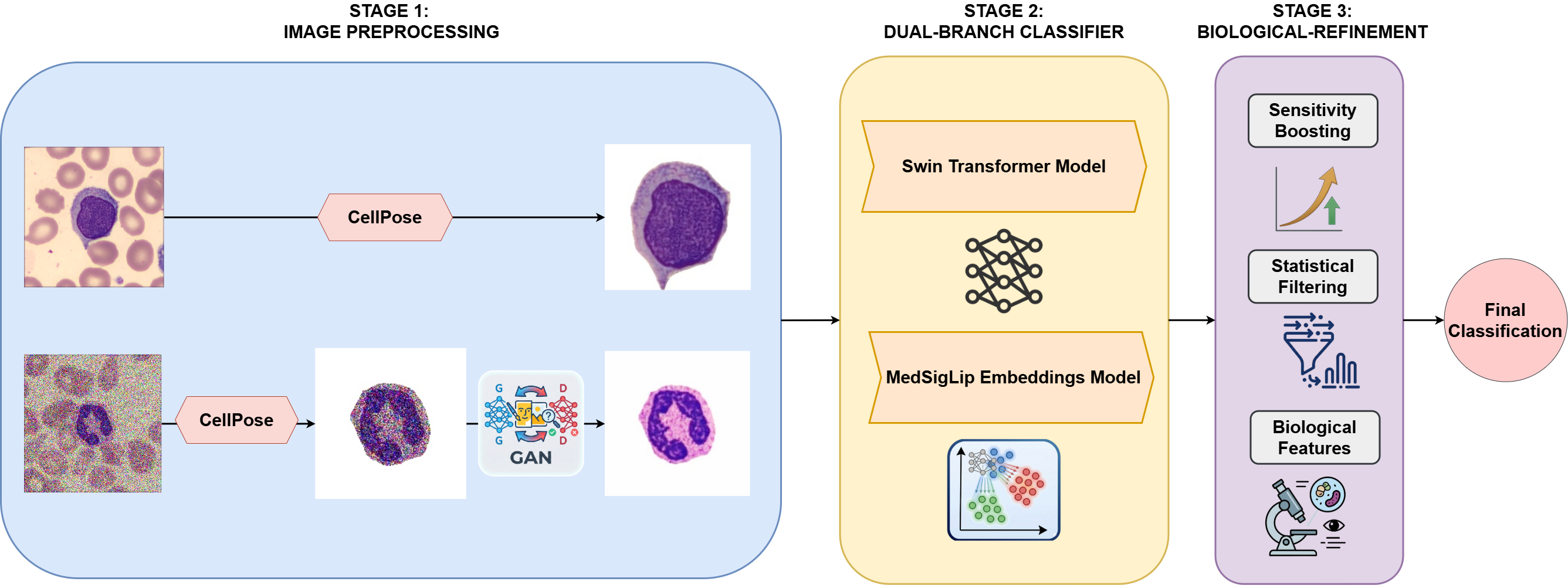}
  \caption{Overview of the proposed three-stage hybrid framework. The pipeline seamlessly integrates generative domain restoration (Stage 1), dual-branch semantic feature extraction (Stage 2), and biological filtering (Stage 3) to achieve robust classification under extreme long-tail distributions.}
  \label{fig:overall_pipeline}
\end{figure*}

Raw microscopy images often contain surrounding red blood cells and staining artifacts. To ensure downstream models focus exclusively on leukocyte morphology, we implement a two-step preprocessing pipeline, as shown in Fig. \ref{fig:overall_pipeline}.

First, we utilize the CellPose \cite{stringer2025cellpose3} framework to accurately localize and isolate the primary WBC, effectively removing irrelevant background context. Second, to address the destructive salt-and-pepper noise introduced by simulated domain shifts, we employ a synthetic paired-training Pix2Pix GAN \cite{isola2017image} restoration strategy. 

As the dataset lacks paired clean–noisy ground truth, we design a custom domain adaptation workflow:
\begin{itemize}
\item \textbf{Data Segregation:} We first apply a quantitative noise filter (based on median blur residuals) to partition the raw dataset into noisy and clean subsets.
    \item \textbf{Synthetic Pair Generation:} We develop a custom noise generator that injects simulated salt-and-pepper noise into the clean subset, thereby generating synthetically paired training samples.
    \item \textbf{GAN Training:} A Pix2Pix GAN architecture is then trained exclusively on these synthetic pairs. The generator learns the complex non-linear mapping required to translate the corrupted domain back into the high-fidelity biological domain.
    \item \textbf{Inference and Restoration:} Finally, the trained GAN is applied to the original noisy subset. This inference effectively restores critical high-frequency morphological details, such as cytoplasmic granules and nuclear textures, ensuring standardized input quality before the feature extraction stage.
\end{itemize}
\subsection{Dual-Branch Feature Classifier}
\label{ssec:extraction}
To capture hierarchical textures and medically aligned semantic features, we propose a parallel dual-branch architecture. Specifically, we independently train two distinct classification models to ensure that the representations remain complementary and non-redundant.

\begin{itemize}
\item \textbf{Hierarchical Transformer Branch \cite{liu2021swin}:} A Swin-T Transformer backbone captures spatial hierarchies. To counteract majority class dominance, this branch is optimized using a cost-sensitive Focal Loss ($\gamma=2.0$) parameterized by inverse log-frequency class weights, alongside a distribution-aware weighted random sampler.
    \item \textbf{Contrastive Branch (MedSigLIP):} We utilize pre-extracted embeddings from MedSigLIP \cite{zhai2023sigmoid}. An advanced projection head maps these embeddings into a lower-dimensional normalized latent space, yielding a secondary, context-rich set of class probabilities $P_{med}$.
\end{itemize}

\subsection{Adaptive Hybrid Refinement Strategy}
\label{ssec:refinement}
Deep models frequently misclassify rare pathological cells as common mature lymphocytes (LY) in out-of-distribution (OOD) scenarios. To recover these suppressed predictions, we implement an adaptive divide-and-conquer strategy (Algorithm \ref{alg:refinement}) that integrates sensitivity boosting, semantic verification, and biological constraints.

Specifically, the biological filtering (Phase 3 of Algorithm \ref{alg:refinement}) addresses two primary morphological confusion pairs:
\begin{itemize}
    \item \textbf{PLY vs. LY (Geometric Spikiness):} Pathological PLY instances often exhibit increased contour irregularity compared to smooth mature LYs. We quantify this geometric distinction via a Spikiness Score $\mathcal{S}$, identifying PLY candidates through statistical outlier thresholds.
    \item \textbf{PC vs. LY/VLY (Mahalanobis Constraints):} Plasma cells possess a voluminous cytoplasm that displaces the nucleus eccentrically. We extract a morphological vector $\mathcal{V}$ (Nuclear-Cytoplasmic ratio, staining intensity, and centroid offset) and apply a Mahalanobis distance filter $D_M$, parameterized by group-specific covariance matrices, to isolate definitive PC samples.
\end{itemize}

\begin{algorithm}[H]
\caption{Suppress-and-Rescue Inference Strategy}
\label{alg:refinement}
\begin{algorithmic}[1]
\State \textbf{Input:} Image $X$, Transformer $M_{Swin}$, MedSigLIP Head $M_{Med}$
\State \textbf{Parameters:} Boost factors $B_c$, Semantic thresholds $\tau$, Morphological thresholds $\tau_{\mathcal{S}}, \tau_{\mathcal{M}}$
\State \textbf{Output:} Final predicted class $\hat{y}$
\State $P_{Swin} \gets M_{Swin}(X)$ \Comment{Base hierarchical probabilities}
\State $y_{base} \gets \arg\max(P_{Swin})$ \Comment{Initial prediction}
\Statex
\State \textbf{Phase 1: Sensitivity Boosting}
\For{$c \in \{\text{PLY}, \text{PC}\}$}
    \State $\hat{P}_{Swin}[c] \gets P_{Swin}[c] \times B_c$ \Comment{Scale by inverse log-frequency}
\EndFor
\State $C_{cand} \gets \arg\max(\hat{P}_{Swin})$
\If{$C_{cand} \notin \{\text{PLY}, \text{PC}\}$}
    \State \Return $y_{base}$ \Comment{No rare candidate surfaced}
\EndIf
\Statex
\State \textbf{Phase 2: Contrastive Verification}
\State $P_{Med} \gets M_{Med}(X)$
\If{$P_{Med}[C_{cand}] < \tau$}
    \State \Return $y_{base}$ \Comment{Does not satisfy semantic verification}
\EndIf
\Statex
\State \textbf{Phase 3: Biological Filtering (Heuristics)}
\If{$C_{cand} == \text{PLY}$}
    \State Compute Spikiness Score $\mathcal{S}(X)$
    \If{$\mathcal{S}(X) > \tau_{\mathcal{S}}$} \Return PLY \Else\ \Return $y_{base}$ \EndIf
\ElsIf{$C_{cand} == \text{PC}$}
    \State Extract feature vector $\mathcal{V}(X)$ via K-Means clustering ($K=2$)
    \State Compute Mahalanobis distance $D_M(\mathcal{V}(X))$
    \If{$D_M \le \tau_{\mathcal{M}}$} \Return PC \Else\ \Return $y_{base}$ \EndIf
\EndIf
\end{algorithmic}
\end{algorithm}
% --- SECTION 3: EXPERIMENTAL RESULTS ---
\section{EXPERIMENTAL RESULTS}
\label{sec:results}

\subsection{Implementation Details and Augmentation}
\label{ssec:setup}
Experiments were conducted in PyTorch on a single NVIDIA H100 80GB GPU. To simulate real-world physical variations in blood smears and prevent overfitting, we applied a comprehensive data augmentation strategy, including spatial transformations (random horizontal/vertical flips, rotations up to $90^\circ$) and color jittering. 

The Swin-T backbone was optimized using the AdamW optimizer (learning rate $1e-4$). We also integrated a log-based \texttt{WeightedRandomSampler} and a cost-sensitive Focal Loss. For the secondary MedSigLIP stage, to maximize the semantic learning of rare cells, the contrastive head was trained on the fully combined dataset over 15 epochs. 

\subsection{Internal Validation}
\label{ssec:internal_val}
Table \ref{tab:internal_metrics} presents the performance of the Swin-T backbone across the 5-fold cross-validation. Note that the MedSigLIP contrastive head is excluded from this internal validation as it was strategically trained on the full dataset to capture a global semantic manifold.

\begin{table}[htb]
\centering
\caption{5-Fold Cross-Validation performance of the Swin-T baseline.}
\label{tab:internal_metrics}
\resizebox{0.85\columnwidth}{!}{%
\begin{tabular}{lc}
\toprule
\textbf{Metric} & \textbf{Mean $\pm$ Std} \\
\midrule
Macro-Averaged F1 & 0.7355 $\pm$ 0.0237 \\
Balanced Accuracy & 0.7226 $\pm$ 0.0279 \\
Macro-Precision & 0.7596 $\pm$ 0.0206 \\
Macro-Specificity & 0.9932 $\pm$ 0.0002 \\
\bottomrule
\end{tabular}%
}
\end{table}

\noindent\textbf{Model Capability Analysis.} The baseline Swin-T model demonstrates consistently high Macro-Specificity ($0.9932$), indicating a strong ability to correctly identify true negatives and avoid false alarms for majority classes. However, the moderate Balanced Accuracy ($0.7226$) and Macro-F1 ($0.7355$) highlight the model's inherent struggle to recall extremely rare pathological cells using purely data-driven features, necessitating our subsequent Suppress-and-Rescue refinement stages.

\subsection{Ablation Study and Leaderboard Performance}
\label{ssec:ablation}
While the baseline model achieved a respectable $0.7355$ Macro-F1 on the internal validation set, submitting a single-fold model to the official hidden test set yielded only $0.63857$ Macro-F1 (Table \ref{tab:ablation}). This substantial performance drop reveals a severe OOD shift between the training data and the evaluation domain. 

To combat the extreme 1,200-fold class imbalance, we initially experimented with established long-tail recognition techniques. As shown in Table \ref{tab:ablation}, standard data-level interventions like Random Over-sampling, as well as state-of-the-art algorithmic approaches including Label-Distribution-Aware Margin Loss (LDAM) \cite{cao2019learning} and Decoupling Representation and Classifier \cite{kang2019decoupling}, were evaluated. However, these purely data-driven methods plateaued at approximately $0.66$ Macro-F1 on the leaderboard. This stagnation highlights a critical limitation: traditional imbalanced learning strategies fail to generalize under the severe morphological overlaps and synthetic domain shifts inherent to the WBCBench dataset.

\begin{table}[htb]
\centering
\caption{Ablation study evaluating conventional long-tail methods versus our proposed framework components on the official private leaderboard.}
\label{tab:ablation}
\resizebox{0.95\columnwidth}{!}{%
\begin{tabular}{lc}
\toprule
\textbf{Method} & \textbf{Macro-F1 (LB)} \\
\midrule
Baseline (Single Fold Swin-T) & 0.63857 \\
Swin-T (5-Fold + TTA) + Over-sampling & 0.65318 \\
Swin-T (5-Fold + TTA) + LDAM Loss \cite{cao2019learning} & 0.66520 \\
Swin-T (5-Fold + TTA) + Decoupling Classifier \cite{kang2019decoupling} & 0.66251 \\
\midrule
Swin-T (5-Fold + TTA) & 0.66585 \\
+ MedSigLIP Semantic Verification (Phase 2) & 0.71528 \\
\textbf{+ Biological Filtering (Phase 3)} & \textbf{0.77139} \\
\bottomrule
\end{tabular}%
}
\end{table}

As detailed in Table \ref{tab:ablation}, applying a 5-fold ensemble with Test-Time Augmentation (TTA) mitigated some variance, matching the state-of-the-art long-tail learning methods ($0.66585$). Breaking through this algorithmic bottleneck required external semantic guidance. The integration of the MedSigLIP contrastive verification (Algorithm \ref{alg:refinement}, Phase 2) provided a substantial boost to $0.71528$. However, the most significant improvement was achieved by introducing the biologically-inspired morphological constraints (Phase 3). By effectively recovering misclassified OOD rare cells that are often suppressed by purely data-driven models, our framework improves the Macro-F1 score to $0.77139$.

% --- SECTION 4: CONCLUSION ---
\section{CONCLUSION}
\label{sec:conclusion}
We presented a robust hybrid framework that synergizes deep learning architectures with biologically-inspired heuristics, achieving a top-tier ranking in the ISBI 2026 WBCBench Challenge. While our current morphological filtering explicitly targets the most critical clinical confusion pairs, future work will focus on generalizing this heuristic approach across all minority leukocyte categories. Furthermore, we plan to more deeply integrate traditional machine learning algorithms with deep semantic embeddings, ultimately creating a comprehensive and clinically reliable diagnostic pipeline for extreme long-tail distributions.
% --- ETHICS & ACKNOWLEDGMENTS ---
\section{COMPLIANCE WITH ETHICAL STANDARDS}
This research was conducted using human subject data made available by the ISBI 2026 WBCBench Challenge. Ethical approval was not required as confirmed by the license attached with the open-access data.

\section{ACKNOWLEDGMENTS}
This work was supported by VinUniversity’s Seed Grant Program under project VUNI.2425.EME.005. The authors would like to thank the VinUni–Illinois Smart Health Center for their continuous support.

\newpage
% --- REFERENCES ---
\bibliographystyle{IEEEbib}
\bibliography{refs}

\end{document}